\documentclass[letterpaper, 10 pt, conference]{ieeeconf} 

\IEEEoverridecommandlockouts          
                   \usepackage{graphicx}
\usepackage{multicol}
\usepackage{cuted}
\usepackage{lipsum}
\let\labelindent\relax
\usepackage{enumitem}
\usepackage{soul}
\usepackage{xspace}
\usepackage{amsmath}
\usepackage{amssymb}
\usepackage{booktabs}
\usepackage{multirow}
\usepackage{array}
\usepackage{makecell}
\usepackage{threeparttable}
\usepackage{adjustbox}
\usepackage[table]{xcolor}
\usepackage{algorithm}
\usepackage{algpseudocode}
\usepackage{caption}
\usepackage{subcaption}
\usepackage{cite}
\usepackage{url}
\usepackage[pagebackref,breaklinks,colorlinks]{hyperref}
\usepackage{cleveref}

\definecolor{customblue}{HTML}{ccf2f5}

\newcommand{\name}[0]{Goal-VLA\xspace} 
\newcommand{\cmarkg}{\textcolor{green!60!black}{$\checkmark$}}
\newcommand{\xmarkr}{\textcolor{red}{$\times$}}
\title{\LARGE \bf \name: Image-Generative VLMs as Object-Centric World Models Empowering Zero-shot Robot Manipulation}
\author{Haonan Chen$^{*1}$, Jingxiang Guo$^{*1}$, Bangjun Wang$^{2}$, Tianrui Zhang$^{1}$, Xuchuan Huang$^{3}$,\\ Boren Zheng$^{4}$, Yiwen Hou$^{1}$, Chenrui Tie$^{1}$, Jiajun Deng$^{1}$, Lin Shao$^{\dagger 1}$%
\thanks{* denotes equal contribution; $\dagger$ denotes the corresponding author.}
\thanks{$^{1}$School of Computing, National University of Singapore; $^{2}$The HKU Musketeers Foundation Institute of Data Science, The University of Hong Kong; $^{3}$Yuanpei College, Peking University; $^{4}$Department of Automation, Tsinghua University.}% <-this % stops a space
}

\begin{document}

\maketitle
\thispagestyle{empty}
\pagestyle{empty}

\begin{strip}
\vspace{-23mm}
  \begin{center}
   \includegraphics[width=1.0\linewidth]{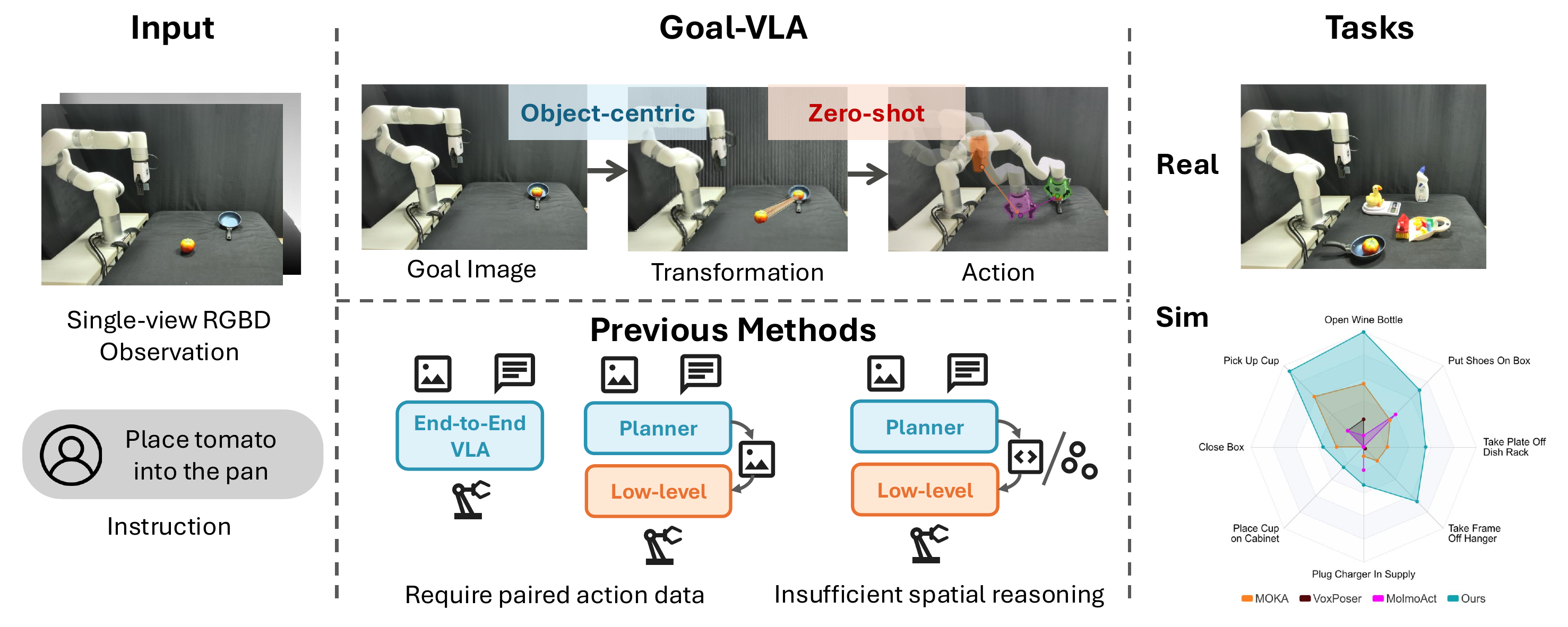}
  \end{center}
  \vspace{-3mm}
  \captionof{figure}{
  \textbf{\name} maps a single-view RGB-D image and a language instruction to executable manipulation actions. Our approach employs an object-centric world model to generate a goal image, from which the corresponding object transformation is subsequently computed, enabling zero-shot manipulation. This overcomes key limitations of previous methods, which often either rely on paired action data for training or lack precise spatial reasoning. We demonstrate our framework's strong performance and generalization capabilities across both simulated and real-world experiments.
  }
  \vspace{-3mm}
\label{fig:teaser}
\end{strip}

%%%%%%%%%%%%%%%%%%%%%%%%%%%%%%%%%%%%%%%%%%%%%%%%%%%%%%%%%%%%%%%%%%%%%%%%%%%%%%%%
\begin{abstract}
Generalization remains a fundamental challenge in robotic manipulation.  
To tackle this challenge, recent Vision-Language-Action (VLA) models build policies on top of Vision-Language Models (VLMs), seeking to transfer their open-world semantic knowledge. However, their zero-shot capability lags significantly behind the base VLMs, as the instruction-vision-action data is too limited to cover diverse scenarios, tasks, and robot embodiments.
In this work, we present \textbf{\name}, a zero-shot framework that leverages Image-Generative VLMs as world models to generate desired goal states, from which the target object pose is derived to enable generalizable manipulation. The key insight is that object state representation is the golden interface, naturally separating a manipulation system into high-level and low-level policies. This representation abstracts away explicit action annotations, allowing the use of highly generalizable VLMs while simultaneously providing spatial cues for training-free low-level control.
To further improve robustness, we introduce a Reflection-through-Synthesis process that iteratively validates and refines the generated goal image before execution.
Both simulated and real-world experiments demonstrate that our \name achieves strong performance and inspiring generalizability in manipulation tasks. Supplementary materials are available at \href{https://nus-lins-lab.github.io/goalvlaweb/}{https://nus-lins-lab.github.io/goalvlaweb/}.
\end{abstract}

\section{Introduction}
\label{Introduction}

Robotic manipulation is a long-standing topic for robotics and embodied AI~\cite{xu2024survey, liu2024aligning, xian2023towards}. It serves as a cornerstone for a wide range of applications in robotics, from household assistance and healthcare to manufacturing and logistics.
Despite recent progress~\cite{shao2025large}, methods that perform exceptionally well in specific settings often collapse under minor distributional shifts~\cite{roy2021machine, kroemer2019review} across environments, tasks, object categories, and robot embodiments. This fact makes robust generalization a valid problem and draws increasing research interest from the community.
This generalization gap is the primary barrier hindering the practical deployment of autonomous robots in unstructured environments.

Foundation models, pre-trained on vast datasets, have emerged as a promising direction to address this challenge. This progress has catalyzed the development of Vision-Language-Action (VLA) models, a broad paradigm aiming to connect multimodal perception with robotic action. Two primary architectural approaches have emerged within the VLA paradigm: end-to-end and hierarchical.

The end-to-end VLA approach is to train a single, large-scale policy to map visual and language inputs to low-level robot actions directly. These models are either developed by finetuning existing Vision-Language Models (VLMs)~\cite{openvla, rdt, molmoact} or by training high-capacity policies from the ground up on robotics data~\cite{rt1, rt2}. While their architectures differ, both approaches share a common and significant challenge: their performance is contingent on massive paired instruction-vision-action data. This data type is exceptionally costly and time-consuming to acquire in sufficient diversity~\cite{yang2025transferring, openXEmbodiment}, which limits its generalization to novel tasks and scenarios.

The hierarchical VLA approach, in contrast, aims to mitigate this intense data dependency. It decouples the problem using a VLM for high-level planning to generate an intermediate representation, guiding a separate low-level policy~\cite{saycan, voxposer, rekep}. Yet, the design of the intermediate representation presents a critical dilemma. Sparse or symbolic representations, such as language descriptions and keypoints~\cite{saycan, liang2023code, driess2023palm, moka}, lack the precise geometric detail required for complex manipulation. In contrast, while dense visual representations like goal images provide rich information~\cite{susie, 3dvla, wu2024ivideogpt, stone2023open}, they typically require the low-level policy to be explicitly trained on paired action data to interpret them, thus forfeiting the zero-shot objective. Other VLM-generated guidance, like value maps, constraints, or reward functions~\cite{voxposer, rekep, iker}, often suffers from significant inaccuracies, stemming from the inherent weakness of current VLMs in precise spatial reasoning.

To ameliorate this problem, we develop our approach based on this key insight: Effective zero-shot generalization in a low-level policy requires explicit spatial guidance. However, the powerful VLMs serving as our high-level goal generator excel at semantic reasoning but perform worse at precise spatial reasoning. To this end, we propose a decoupled architecture that leverages the VLM as an object-centric world model. Unlike traditional agent-centric world models~\cite{dreamer, wu2023daydreamer, zhu2025unified, wu2024ivideogpt} that are tied to specific robot kinematics and thus limit cross-embodiment generalization, our world model focuses exclusively on the semantic goal in the image space. This allows a separate, dedicated spatial grounding module to handle the translation of this semantic goal into explicit spatial guidance of the object, playing to the strengths of each component.

Formally, we introduce \name, a zero-shot manipulation framework that employs an Image-Generative VLM to produce a goal image depicting the desired task outcome, functioning as an object-centric world model. This visual representation is translated into a precise target pose to direct a low-level policy. To evaluate and optimize the generated goal image, we incorporate a Reflection-through-Synthesis mechanism that assesses and iteratively refines the visual output. The target object state is then extracted from the refined goal image, and the goal pose is computed using feature matching and point cloud registration. This object-centric representation is inherently generalizable across various tasks, environments, object categories, and robot embodiments. By providing an explicit object pose rich in spatial information, our framework effectively guides a low-level policy to perform manipulation tasks in a zero-shot manner.

Our framework operates solely on a user's task description and a single-view RGB-D input, requiring no prior scene knowledge such as pre-scanned maps or object meshes. We demonstrate its effectiveness on tasks in simulated and real-world settings, including \textit{Pick and Place, Table Sweeping, Bottle Stand-Up, and Box Closing}.
Our approach consistently outperforms baseline methods, such as MOKA~\cite{moka}, VoxPoser~\cite{voxposer}, and MolmoAct~\cite{molmoact}, achieving higher success rates. Moreover, our pipeline does not require task-specific fine-tuning, highlighting its zero-shot generalization.

To summarize, our key contributions are:
\begin{itemize}
    \item We introduce \name, a decoupled hierarchical framework that leverages an Image-Generative VLM as a world model to generate goal object states, serving as the bridge between high-level semantic reasoning and low-level action control.
    \item We propose an iterative refinement process, termed Reflection-through-Synthesis, where the generated goal is improved by synthesizing and overlaying a virtual image of the target object onto the current observation for visual inspection and improvement.
    \item We demonstrate that \name achieves inspiring zero-shot generalizability across diverse manipulation tasks, environments, object categories, and robot embodiments in both simulation and the real world, without requiring any task-specific finetuning.
\end{itemize}
\section{Related Work}
\label{Related Work}

\label{Foundation Models for Robotic Manipulation}

\begin{table}[h!]
\vspace{-0.2cm}
\captionsetup{font={large}}
\centering
\resizebox{\columnwidth}{!}{%
\begin{threeparttable}
\caption{Comparison of VLA Paradigms}
\label{tab:vla_comparison_aligned}
\begin{tabular}{@{} 
    >{\centering\arraybackslash}m{1.4cm} 
    >{\centering\arraybackslash}m{1.6cm} 
    >{\centering\arraybackslash}m{1.4cm} 
    >{\centering\arraybackslash}m{1.4cm} 
    >{\centering\arraybackslash}m{1.4cm} 
    >{\centering\arraybackslash}m{1.4cm} @{}}
\toprule
% --- MODIFIED HEADERS HERE FOR ALIGNMENT ---
\textbf{\makecell{Category}} & \textbf{\makecell{Method}} & 
\textbf{\makecell{Future \\[-3pt] State \\[-3pt] Awareness}} & 
\textbf{\makecell{Inter- \\[-3pt] mediate \\[-3pt] Type}} & 
\textbf{\makecell{Training- \\[-3pt] Free \\[-3pt] Policy}} & 
\textbf{\makecell{Precise 3D \\[-3pt] Spatial \\[-3pt] Grounding}} \\
\midrule
\multirow{2}{*}{\textbf{\makecell{End-to-End \\ VLA}}} 
 & \makecell[l]{RT-1/2~\cite{rt1, rt2}, \\ OpenVLA~\cite{openvla}, \\ $\pi_0$~\cite{black2410pi0}, \\ RDT-1B~\cite{rdt}} & \xmarkr & N/A & \xmarkr & \xmarkr \\
 \cmidrule[0.2pt](lr){2-6}
 & MolmoAct~\cite{molmoact} & \xmarkr & Depth \& Trajectory & \xmarkr & \cmarkg \\
\midrule
\multirow{9}{*}{\textbf{\makecell{Hierarchical \\ VLA}}} 
 & \makecell[l]{SayCan~\cite{saycan}, \\  PaLM-E~\cite{driess2023palm}} & \xmarkr & Symbolic Skills & \xmarkr & \xmarkr \\
 \cmidrule[0.2pt](lr){2-6}
 & Code as Policies~\cite{liang2023code} & \xmarkr & Code & \xmarkr & \xmarkr \\
 \cmidrule[0.2pt](lr){2-6}
 & ProgPrompt~\cite{singh2022progprompt} & \xmarkr & Programmatic Plan & \xmarkr & \xmarkr \\
 \cmidrule[0.2pt](lr){2-6}
 & \makecell[l]{MOKA~\cite{moka}} & \cmarkg & Keypoints & \cmarkg & \xmarkr \\
 \cmidrule[0.2pt](lr){2-6}
 & \makecell[l]{Rekep~\cite{rekep}} & \xmarkr & Keypoints & \cmarkg & \xmarkr \\
 \cmidrule[0.2pt](lr){2-6}
 & \makecell[l]{IKER~\cite{iker}} & \cmarkg  & Rewards & \xmarkr & \xmarkr \\
 \cmidrule[0.2pt](lr){2-6}
 & \makecell[l]{SUSIE~\cite{susie}} & \cmarkg & Subgoal Images & \xmarkr & \xmarkr \\
 \cmidrule[0.2pt](lr){2-6}
 & VoxPoser~\cite{voxposer} & \xmarkr & 3D Value Maps & \cmarkg & \xmarkr \\
 \cmidrule[0.2pt](lr){2-6}
 & 3D-VLA~\cite{3dvla} & \cmarkg & 3D Scene & \xmarkr & \cmarkg \\
 \cmidrule[0.2pt](lr){2-6}
 &\textbf{\name (Ours)} & \cmarkg & Object State & \cmarkg & \cmarkg \\
\bottomrule
\end{tabular}
\end{threeparttable}
} 
\end{table}

\subsection{Foundation Models Paradigms for Robotic Manipulation}
Recent advances in foundation models have significantly influenced robotics, especially in robotic manipulation tasks, by leveraging LLMs and VLMs for high-level planning and decision-making~\cite{firoozi2025foundation, ma2024survey, manual2skill, zhong2025survey}. A key development in this area is the Vision-Language-Action (VLA) model: a class of frameworks that maps multimodal inputs, typically visual observations and language instructions, to executable robot actions~\cite{ma2024survey, zhong2025survey}. Current research building on this concept can be broadly categorized into several main paradigms, 
which we compare in detail in Table~\ref{tab:vla_comparison_aligned}.

\textbf{End-to-End Vision-Language-Action Models.}
The first paradigm aims to build monolithic, end-to-end models that directly map visual and linguistic inputs to low-level robot actions. This approach includes methods that train large models from scratch, such as RT-1~\cite{rt1} and RT-2~\cite{rt2}, as well as those that finetune pre-trained VLMs, like OpenVLA~\cite{openvla}, RDT-1B~\cite{rdt}, and MolmoAct~\cite{molmoact}. While these models have impressive capabilities, their performance is fundamentally bottlenecked by the need for massive datasets of paired instruction-vision-action data. The high cost and complexity of collecting such data severely constrain their diversity, limiting the models' zero-shot generalization.

\textbf{Hierarchical Models with Intermediate Representations.} A second representative paradigm employs foundation models in a hierarchical structure. A high-level VLM is a planner in this setup, generating an intermediate representation to guide a separate low-level policy~\cite{yao2023react}. Therefore, the design of this intermediate representation is critical, defining the primary trade-offs within this paradigm. One line of work utilizes symbolic or sparse geometric representations. This includes selecting from a library of predefined skills~\cite{saycan, driess2023palm}, generating executable code~\cite{singh2022progprompt, liang2023code}, or defining tasks through a set of keypoints~\cite{moka, mikami2024natural}. While effective for structured tasks, the generalizability of these methods is often constrained by the limited scope of discrete skill sets or the insufficiency of sparse keypoints to capture complex object interactions. Other methods generate more explicit, dense visual guidance to provide richer context, such as subgoal images~\cite{susie, 3dvla} or videos~\cite{gen2act, wu2024ivideogpt}. A significant drawback, however, is that the low-level policy typically needs to be trained to interpret these visual goals, which reintroduces a data dependency and undermines the objective of a truly zero-shot, training-free system. A third approach involves the VLM generating implicit spatial guidance, like value maps~\cite{voxposer}, constraints~\cite{rekep}, or reward functions~\cite{iker}. These methods require the VLM to ground its reasoning in 3D space, a capability that is often imprecise due to the inherent limitations of its spatial understanding.

\subsection{Reflection in Foundation Models}

Reflection mechanisms, which enable generative models to iteratively critique and refine their outputs, have recently attracted growing attention as a promising approach for enhancing robotic manipulation capabilities. In generative modeling, recent studies demonstrate that incorporating self-feedback or iterative critiques substantially improves the quality and coherence of generated outputs~\cite{shinn2023reflexion,madaan2023self,raman2024cape}. Notable examples include CritiqueLLM~\cite{ke2023critiquellm} and Idea2Img~\cite{yang2024idea2img}, which showcase how reflective feedback loops facilitate progressive refinement and correction of initial predictions. Extending these reflective approaches into robotics, several recent frameworks~\cite{wang2023describe, feng2025reflective,yang2024guiding} integrate self-reflection into robotic task planning and action execution, enabling robotic agents to dynamically identify and correct errors, thereby progressively enhancing their performance during tasks. Moreover, additional studies have advanced this concept by incorporating multimodal reflection mechanisms, effectively bridging high-level cognitive reasoning with low-level motor control adjustments. This multimodal integration significantly improves robot robustness and adaptability, enabling robots to manage uncertainties better and effectively generalize across diverse manipulation scenarios and real-world conditions~\cite{li2025selfcorrectingvisionlanguageactionmodelfast,xiong2024aic,xia2025phoenix}.
\section{Method}
\label{Method}

\begin{algorithm}[b!]
\caption{\name Execution Framework}
\label{alg:main_algorithm}
\begin{algorithmic}[1]
    \Require Initial observation $O = (\mathcal{I}, \mathcal{D})$, Language instruction $\mathcal{L}$, Initial End-effector pose $P_{init}$
    \Ensure Action sequence $\{\boldsymbol{a}\}_i$

    \Procedure{\name}{$O, \mathcal{L}$}
        \Statex \textbf{Stage 1: Goal State Reasoning}
        \State $\mathcal{L}_{e} \gets \text{EnhancePrompt}(\mathcal{I}, \mathcal{L})$
        
        \Statex \# \textit{--- Reflection-through-Synthesis Loop ---}
        \For{$i=1$ \textbf{to} max\_iterations}
            \State $\mathcal{I}'_{cand} \gets \text{GenerateImage}(\mathcal{I}, \mathcal{L}_{e})$
            \State $\mathcal{I}_{synth} \gets \text{SynthesizeOverlay}(\mathcal{I}, \mathcal{I}'_{cand})$
            \State is\_valid, $\mathcal{L}_{revised} \gets \text{Reflector}(\mathcal{I}_{synth}, \mathcal{L}_{e})$
            \If{is\_valid} \textbf{break} \EndIf
            \State $\mathcal{L}_{e} \gets \mathcal{L}_{revised}$
        \EndFor
        \State $\mathcal{I}', \mathcal{M}, \mathcal{M}', \mathcal{D}' \gets \text{GenMask\&Depth}( \mathcal{I}, \mathcal{I}'_{cand})$

        \Statex \textbf{Stage 2: Spatial Grounding}
        \State $(R, t) \gets \text{ComputeTransform}(\mathcal{I}, \mathcal{I}', \mathcal{D}, \mathcal{D}', \mathcal{M}, \mathcal{M}')$

        \Statex \textbf{Stage 3: Low-level Policy}
        \State $P_{contact} \gets \text{DetermineContactPose}(O, \mathcal{M})$
        \State $P_{goal} \gets \text{ApplyTransform}((R, t), P_{contact})$
        \State $\{\boldsymbol{a}\}_i \gets \text{MotionPlan}(P_{init}, P_{contact}, P_{goal})$
        \State \Return $\{\boldsymbol{a}\}_i$
    \EndProcedure
\end{algorithmic}
\end{algorithm}

\begin{figure*}[!ht]
  \begin{center}
   \includegraphics[width=1.0\linewidth]{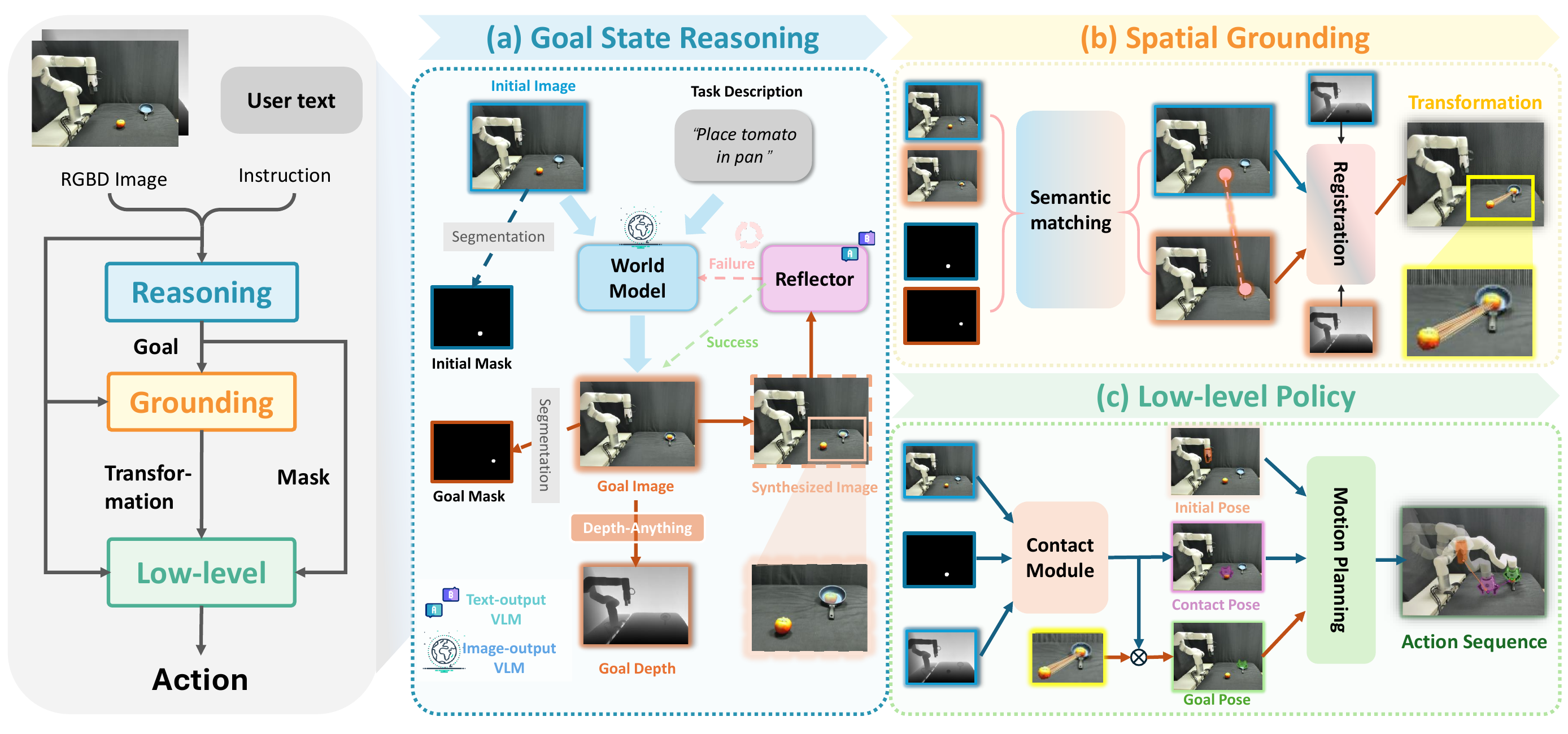}
  \end{center}
      \caption{%\honam{Change Overview(Fonts, Lines(topleft), Size)}
      Overview of the \textbf{\name} framework, which decouples the manipulation pipeline into three stages:
      \textbf{(a) Goal State Reasoning:} A VLM generates a goal image from instructions and refines it for task feasibility, yielding a validated goal with image, mask, and depth.
      \textbf{(b) Spatial Grounding:} The object's transformation is computed by feature matching and point cloud registration between the initial and goal states.
      \textbf{(c) Low-level Policy:} The gripper's goal pose is derived by applying the object's transformation to a contact pose, after which a motion planner generates the final trajectory for robot execution.}
\label{fig:overview}
\end{figure*}

The overall workflow of our framework is illustrated conceptually in Figure~\ref{fig:overview} and detailed procedurally in Algorithm~\ref{alg:main_algorithm}. This section begins by formulating the problem and defining the inputs and outputs for each module of our hierarchical pipeline (Sec. ~\ref{Problem Formulation}). We then detail the core components, starting with the Goal State Reasoning module, which interprets the user's instruction to generate a visual goal state (Sec. ~\ref{Goal State Reasoning}). Subsequently, the Spatial Grounding module takes this visual representation and computes a precise 3D transformation (Sec. ~\ref{Spatial Grounding}). Finally, we describe how the Low-level Policy uses this transformation to plan and execute the physical manipulation task (Sec. ~\ref{Low-level Policy}).

\subsection{Problem Formulation}
\label{Problem Formulation}

Given a single-view RGBD image observation $O = (\mathcal{I} \in \mathbb{R}^{H \times W \times 3}, \mathcal{D} \in \mathbb{R}^{H \times W \times 1})$, and a natural language task description \(\mathcal{L}\) , the objective is to generate an action sequence $\{\boldsymbol{a}\}_i$ that completes the manipulation task described in $\mathcal{L}$. 

The \textbf{Goal State Reasoning} receives an RGB image \(\mathcal{I}\) and a task description \(\mathcal{L}\) to produce a goal image \(\mathcal{I}' \in \mathbb{R}^{H \times W \times 3}\) and a goal metric depth map \(\mathcal{D}' \in \mathbb{R}^{H \times W \times 1}\). The Goal State Reasoning module also generates the object masks for the initial and goal images, denoted as $\mathcal{M}$ and $\mathcal{M'}$.

The \textbf{Spatial Grounding} is given with the initial and goal RGB images $\mathcal{I}, \mathcal{I}'$, real depth maps $\mathcal{D}$ and goal metric depth $\mathcal{D}'$, and object masks $\mathcal{M}, \mathcal{M}'$. The objective of this module is to compute the rotation $R\in SO(3)$ and translation $t\in \mathbb{R}^3$ that maps the object from its initial pose to its goal pose.

The \textbf{Low-level Policy} takes the current observation $\mathcal{O} = (\mathcal{I}, \mathcal{D})$ and the mask $\mathcal{M}$ as input, then outputs a sequence of actions $\{\boldsymbol{a}\}_i$ to drive the system from $\mathcal{O}$ to $\mathcal{O}'=(\mathcal{I'}, D')$.

\subsection{Goal State Reasoning}
\label{Goal State Reasoning}

The Goal State Reasoning module is responsible for translating the user's abstract language instruction ($\mathcal{L}$) into a concrete and plausible visual goal, represented by a goal image ($\mathcal{I}'$) and a goal depth map ($\mathcal{D}'$). This is achieved through an initial prompt enhancement step and an iterative refinement loop we term Reflection-through-Synthesis.

Before the image generation process, we enhance the user's raw instruction to make it more descriptive for the image generation model. The initial image $\mathcal{I}$ and the raw task description $\mathcal{L}$ are fed into a text-output VLM (e.g., Gemini 2.5 Pro). This model enriches the concise instruction with critical details inferred from the visual context, producing a more descriptive enhanced prompt.

\textbf{The Reflection-through-Synthesis Loop:} This enhanced prompt then seeds our iterative refinement loop, which generates and validates candidate goals until one is deemed successful. Each iteration involves the following steps:
\begin{itemize}
    \item \textbf{Goal Image Generation:} The current prompt and the initial image $\mathcal{I}$ are passed to a pre-trained image generation VLM (Gemini 2.5 Flash-image, also known as Nano Banana) to produce a candidate goal image.
    \item \textbf{Synthesis for Inspection:} We construct a synthesized image to provide the Reflector VLM with a clear basis for evaluation.  Using Grounded SAM~\cite{groundedsam}, we segment the target object from the candidate goal and overlay this ``virtual image'' onto the initial scene with partial transparency. This overlay is crucial as it provides an in-context visualization of the goal, which mitigates the semantic gap and enables a more robust evaluation.
    \item \textbf{Reflection and Refinement:} The synthesized image, along with the original instruction and the current enhanced prompt, is fed to a \textbf{Reflector VLM} (Gemini 2.5 Pro). The Reflector assesses if the synthesized outcome aligns with the task's semantic requirements. If successful, the loop terminates. If not, the Reflector generates a revised prompt containing corrective feedback, which is used for the next generation iteration.
\end{itemize}

This process continues until the Reflector validates a goal image or a pre-set maximum number of attempts is reached. Once a goal image is accepted, the final outputs for the next module are generated: a depth estimation model (Depth-Anything V2~\cite{yang2024depth}) produces the goal's relative depth map. At the same time, Grounded SAM~\cite{groundedsam} extracts the target object masks for both the initial and final goal images.

\subsection{Spatial Grounding}
\label{Spatial Grounding}
%To determine the spatial transformation between the initial scene and the generated goal, we first establish pixel-level correspondences and then register the corresponding 3D point clouds. 
The Spatial Grounding module consists of two main stages: \textbf{semantic matching} and \textbf{point-cloud registration}.

\textbf{Semantic Matching.} Given the initial observation $\mathcal{I}\in\mathbb{R}^{H\times W\times 3}$ and the goal image $\mathcal{I}'\in\mathbb{R}^{H\times W\times 3}$, our goal is to obtain a pixel matching $\mathcal{F}: (x, y)\to (x', y')$, which maps pixel in the initial image $\mathcal{I}$ to the corresponding pixel in the goal image $\mathcal{I}'$. Since the goal image synthesized by the generative VLM is semantically correct but may not preserve the instance-level appearance of the original object, traditional methods like optical flow estimation or 2D pixel tracking yield unreliable results. Therefore, we utilize semantic features to generate robust pixel matching. Based on the existing semantic matching model Geo-Aware~\cite{zhang2024telling}, we extract pixel-wise semantic features $\mathbf{f}, \mathbf{f'}$ from $\mathcal{I}$, $\mathcal{I}'$, respectively. Furthermore, we match pixels between the two frames by pairing the pixel in $\mathcal{I}$ with the pixel yielding the highest cosine similarity among all pixels in $\mathcal{I}'$, i.e.
\begin{equation}
    (x', y')=\underset{(p,q)}{\arg\max}\frac{\mathbf{f}_{(x,y)}\cdot \mathbf{f}'_{(p,q)}}{\|\mathbf{f}_{(x,y)}\|\|\mathbf{f}'_{(p,q)}\|}.
\end{equation}

% Furthermore, since the world model may generate an image $\mathcal{I}'$ containing both the original and target objects, we implement a spatial filter to address the limitations of segmentation tools like GroundedSAM2~\cite{groundedsam}, which struggle to isolate the target. For a pair of corresponding pixels, $(x,y)$ in the original frame and $(x',y')$ in the generated frame, we accept it as a valid match if and only if its displacement exceeds a predefined threshold $\delta$:
% \[
% (x-x')^2+(y-y')^2 \geq \delta.
% \]

\textbf{Point-cloud Registration.} Following the establishment of 2D correspondences, this stage computes the 3D spatial transformation. The process involves lifting the 2D images to 3D point clouds using depth information. While a ground-truth depth map $\mathcal{D}$ is available for the initial frame, the depth map for the generated goal, $\mathcal{D}'$, is estimated using the DepthAnythingV2 model.

To align these different scales, we first perform depth alignment. A linear transformation, characterized by a scale factor $s_1$ and a bias $b$, is calibrated to map the predicted depth scale of $\mathcal{D}'$ to the metric scale of $\mathcal{D}$. This transformation is determined via the method of least squares:
\begin{equation}
    \mathcal{D}[(\mathcal M\cup\mathcal{M}')^c]=s_1\cdot\mathcal{D}'[(\mathcal M\cup\mathcal{M}')^c]+b,
\end{equation}
The regression is performed exclusively on background pixels, denoted by the complement set $(\mathcal M\cup\mathcal{M}')^c$, to ensure manipulated foreground objects do not skew the alignment. The learned parameters are then applied to the entire predicted depth map to yield a correctly scaled version,
\begin{equation}
    \mathcal{D}'_{\rm rescaled}=s_1\cdot D'+b.
\end{equation}
Subsequently, object-specific point clouds, $\mathcal{P}$ and $\mathcal{P}'$, are extracted from the initial and goal frames by applying object masks $\mathcal{M}$ and $\mathcal{M}'$ to the depth maps $\mathcal{D}$ and $\mathcal{D}_{\rm rescaled}'$, respectively. Specifically, we use the pixel matching $\mathcal{F}$ to establish a set of corresponding point pairs, which are then filtered to remove those with negligible changes in distance. To robustly account for scale differences in the generated goal,  we solve for a similarity transformation between the two point clouds:
\begin{equation}
    s_2\cdot\mathcal{P}'=R\mathcal{P}+t,
\end{equation}
where $R$ stands for the rotation; $t$ stands for translation and $s_2$ is the scale factor of the transformation, respectively. The optimal scale $s_2$, rotation $R$ and translation $t$ are computed using the Umeyama algorithm~\cite{umeyama2002least}. These final parameters, representing the precise spatial transformation, are then utilized by a low-level policy module to guide robot actions.

\subsection{Low-level Policy}
\label{Low-level Policy}

The \textbf{Low-level Policy} translates the object goal pose, provided by the high-level modules, into executable robot actions. This process involves three stages: determining an initial contact point on the object, computing the gripper's final goal pose, and planning a valid trajectory. A key advantage of our framework's object-centric representation is its flexibility; it provides an explicit geometric target and is compatible with various low-level controllers operating on different input modalities (e.g., RGB images or point clouds).

\textbf{Contact Module.} The Contact Module determines a feasible contact pose using a multi-stage, sample-based method on the object's point cloud. First, a large set of contact pose candidates is generated on the object's surface, oriented along local surface normals. This set is then pruned through filters that check for valid orientation and perform collision checks using a robot's end-effector geometry against the scene. Finally, the remaining collision-free candidates are scored based on task-relevant geometric heuristics, and the top-scoring pose is selected as the initial contact pose for subsequent planning.

\textbf{Goal Pose Computing.} We assume that the relative pose between the gripper and the object remains constant after contact. This assumption holds for most prehensile actions and a subset of non-prehensile tasks. Based on this, we calculate the gripper's goal pose by applying the object's transformation, as determined by the Spatial Grounding module, to the gripper's initial pose. This provides the target configuration for motion planning.

\textbf{Motion Planning Module.}
Finally, the Motion Planning Module generates a collision-free trajectory for the manipulator from its current configuration to the target pose. In our simulated experiments, we leverage a built-in sample-based planner~\cite{rlbench} to interpolate the trajectory. We utilized a sample-based motion planner in the real-world experiment to generate a feasible path in the robot's configuration space.
\section{Experiment}
\label{Experiments}

In this section, we conduct comprehensive experiments and analyses to answer the following key questions:

\begin{enumerate}[label=Q\arabic*:, leftmargin=2em, labelwidth=1.5em, itemindent=!, align=left]
    \item How well does our proposed method perform compared to existing baselines?
    \item How effective is the Reflection-through-Synthesis process in refining the outputs of the high-level reasoning?
    \item Can our framework generalize across diverse environments, tasks, object categories, and robot embodiments?
\end{enumerate}

\begin{table*}[t!]
  \centering
  \setlength\tabcolsep{6pt} 
  \renewcommand{\arraystretch}{1.2} 
  \begin{threeparttable}
    \captionsetup{width=\linewidth}
    \caption{Results of Simulation Experiments. All methods are evaluated in a zero-shot setting.}
    \label{tab:simulation}
    \begin{tabular}{lccccccccc}
      \toprule
      Method      & \makecell{Pick Up\\Cup} & \makecell{Open Wine\\Bottle} & \makecell{Put Shoe\\On Box} & \makecell{Take Plate\\Off Dish Rack} & \makecell{Take Frame\\Off Hanger} & \makecell{Plug Charger\\In Supply} & \makecell{Place Cup\\on Cabinet} & \makecell{Close\\Box} & \makecell{Average\\Success Rate} \\
      \midrule
      OpenVLA     & 0/100 & 2/100 & 0/100 & 0/100 & 0/100 & 0/100 & 0/100 & 0/100 & 0.2\% \\
      Pi0         & 0/100 & 0/100 & 0/100 & 0/100 & 0/100 & 0/100 & 0/100 & 0/100 & 0.0\% \\
      SUSIE       & 0/100 & 0/100 & 0/100 & 0/100 & 0/100 & 0/100 & 0/100 & 0/100 & 0.0\% \\
      MOKA        & 62/100 & 55/100 & 33/100 & 21/100 & 17/100 & 8/100 & 0/100 & 24/100 & 26.0\% \\
      VoxPoser    & 20/100 & 24/100 & 0/100 & 0/100 & 2/100 & 0/100 & 0/100 & 0/100 & 5.8\% \\
      MomolAct    & 20/100 & 10/100 & 40/100 & 0/100 & 0/100 & 20/100 & 0/100 & 0/100 & 11.3\% \\
      \rowcolor{customblue} \textbf{Ours} & 93/100 & 100/100 & 70/100 & 55/100 & 67/100 & 33/100 & 25/100 & 36/100 & 59.9\% \\
      \bottomrule
    \end{tabular}
  \end{threeparttable}
  \vspace{-2mm}  
\end{table*}

\subsection{Experimental Setup}
\label{Setup}

\textbf{Simulation.} Our simulation experiments are conducted in RLBench~\cite{rlbench}, utilizing a Franka Emika Panda 7-DoF arm with attached RGB-D cameras. The robot arm is fixed to a tabletop, and for each task, objects are placed in randomized initial configurations determined by a random seed. We select eight representative tasks from the RLBench benchmark to evaluate a range of manipulation skills, with the complete list and setup details provided in Table~\ref{tab:simulation}. In all experiments, the robot starts without holding any object at the beginning of each trial.

To robustly assess performance and account for variations in object placement, each task is evaluated across 10 random seeds. For each seed, which defines a unique initial scene arrangement, we conduct 10 independent trials of each method, resulting in 100 evaluation runs per task for calculating the success rate.

\textbf{Real World.} Our real-world experimental setup comprises a 7-DoF UFACTORY X-ARM 7 and an Orbbec Femto Bolt RGB-D camera. We design four distinct real-world tasks to evaluate a range of core manipulation capabilities: \textit{Place Tomato in Pan}, a foundational pick-and-place task requiring reasoning about object containment; \textit{Table Sweeping}, which involves tool use and non-prehensile manipulation; \textit{Weigh the Duck}, a task demanding precise placement onto a small target; and \textit{Stand a Bottle Upright}, which tests object reorientation. We conduct 10 trials for each task, with detailed results presented in Table~\ref{tab:real-world} and Figure~\ref{fig:real world}.

For detailed descriptions of our simulation and real-world tasks, please refer to the Appendix~\ref{Simulation Task Description} and \ref{Real World Task Description}.

\subsection{Baselines}

We evaluate our method against representative baselines from the two primary paradigms discussed in our related work: end-to-end and hierarchical. Our selection is designed to rigorously test our approach against the main competing strategies in robotic manipulation.

For the end-to-end paradigm, we select OpenVLA\cite{openvla}, Pi0\cite{black2410pi0}, and MolmoAct\cite{molmoact}. These models are typical examples of the end-to-end VLA approach that directly maps multimodal inputs to robot actions.

We choose a diverse set of methods for the hierarchical paradigm to evaluate the efficacy of different intermediate representations. Our selection includes: VoxPoser\cite{voxposer} (value maps), SUSIE\cite{susie} (subgoal images), and MOKA\cite{moka} (keypoints), covering a broad spectrum of representations, from implicit spatial guidance to explicit visual targets.

All of our comparative evaluations are conducted under a strict zero-shot protocol, which means neither our method nor the baselines undergo any task-specific fine-tuning.

\subsection{Simulation Experiments (Q1)}

The results of our simulation experiments are presented in Table~\ref{tab:simulation}. Our method, \textbf{\name}, achieves a remarkable average success rate of 59.9\%, significantly outperforming all baselines across a diverse set of eight manipulation tasks.

The end-to-end baselines, OpenVLA and Pi0, fail almost completely (0.2\% and 0\% success rates, respectively), highlighting their brittleness in this zero-shot setting due to a strong dependency on large-scale, in-domain action data. Performance within the hierarchical paradigm was varied but limited by the intermediate representation. While methods using keypoints (MOKA, 26.0\%) or trajectories (MomolAct, 11.3\%) show some capability, their performance lags significantly behind that of our approach. This suggests a fundamental weakness in the baselines' intermediate representations. Unlike our explicit 3D pose, their representations either lack sufficient spatial precision or necessitate task-specific training for the low-level policy.

These findings decisively answer our first research question (Q1). While other methods falter due to data dependency or imprecise goals, \name's explicit goal object state provides the necessary precision and robustness for a low-level policy to succeed.

\subsection{Ablation Study (Q2)}
We perform an ablation study to validate the contributions of our two key components: Input Enhancement and the Reflection-through-Synthesis process.

\begin{figure}[b!]
    \vspace{-5pt}
    \centering
    \includegraphics[width=\columnwidth]{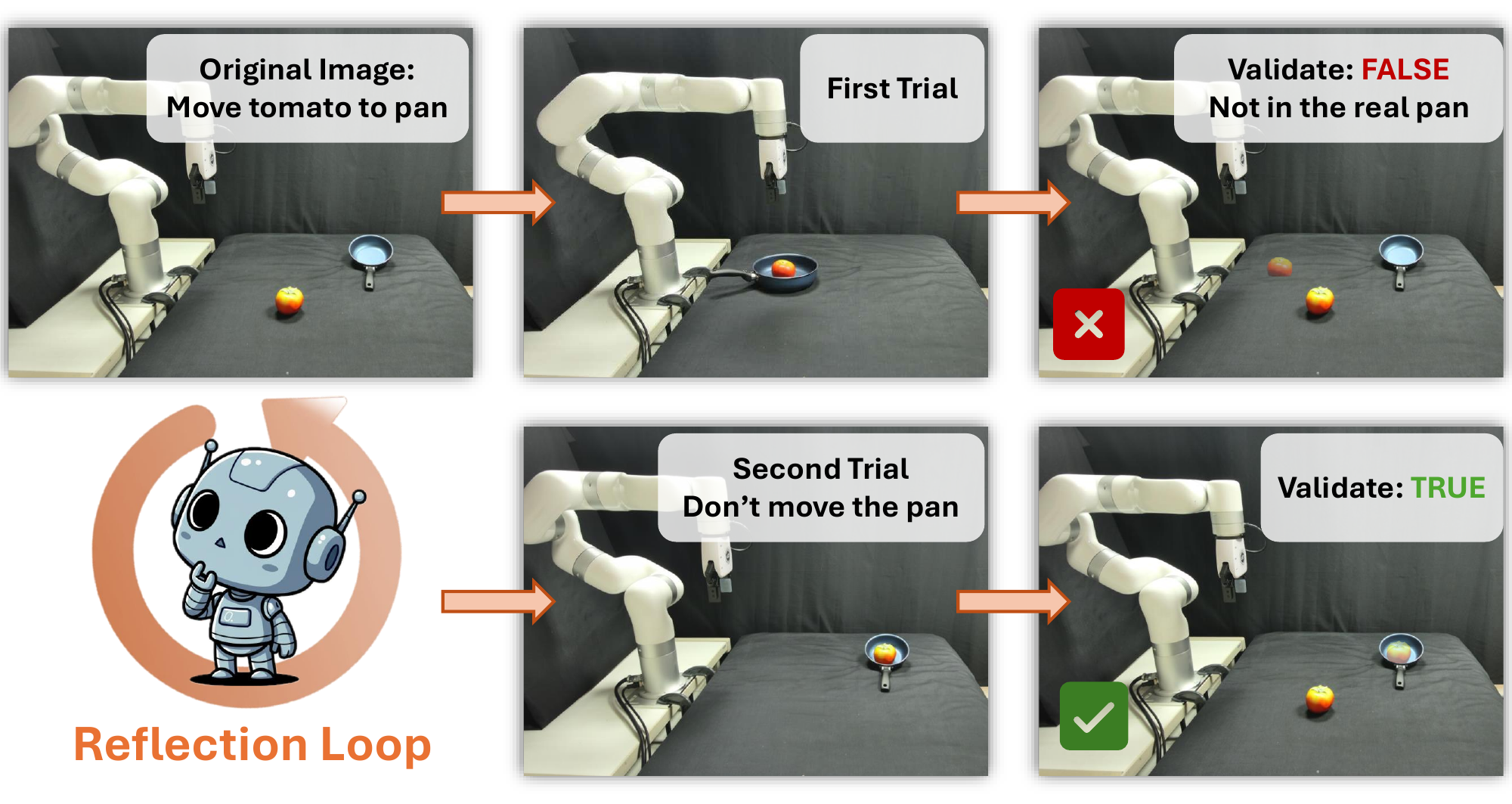}
    \caption{An example of our \textbf{Reflection-through-Synthesis} process, which corrects a semantically correct but infeasible goal by refining the generation prompt.}
    % \vspace{-10pt}
    \label{fig:reflector_workflow}
\end{figure}
% \vspace{-0.8cm}

\textbf{Reflection's Necessary:} Figure~\ref{fig:reflector_workflow} highlights a typical failure mode of image generation. Image-Generative VLM may alter non-target elements (e.g., moving the pan along with the tomato), creating a semantically plausible but physically incorrect goal. Our \textbf{synthesis} step, overlaying the segmented target object onto the original scene, grounds the reflection in the world context. This allows our Reflector VLM to identify such errors and provide corrective feedback effectively.

\begin{figure}[!t]
    \centering
    \includegraphics[width=\columnwidth]{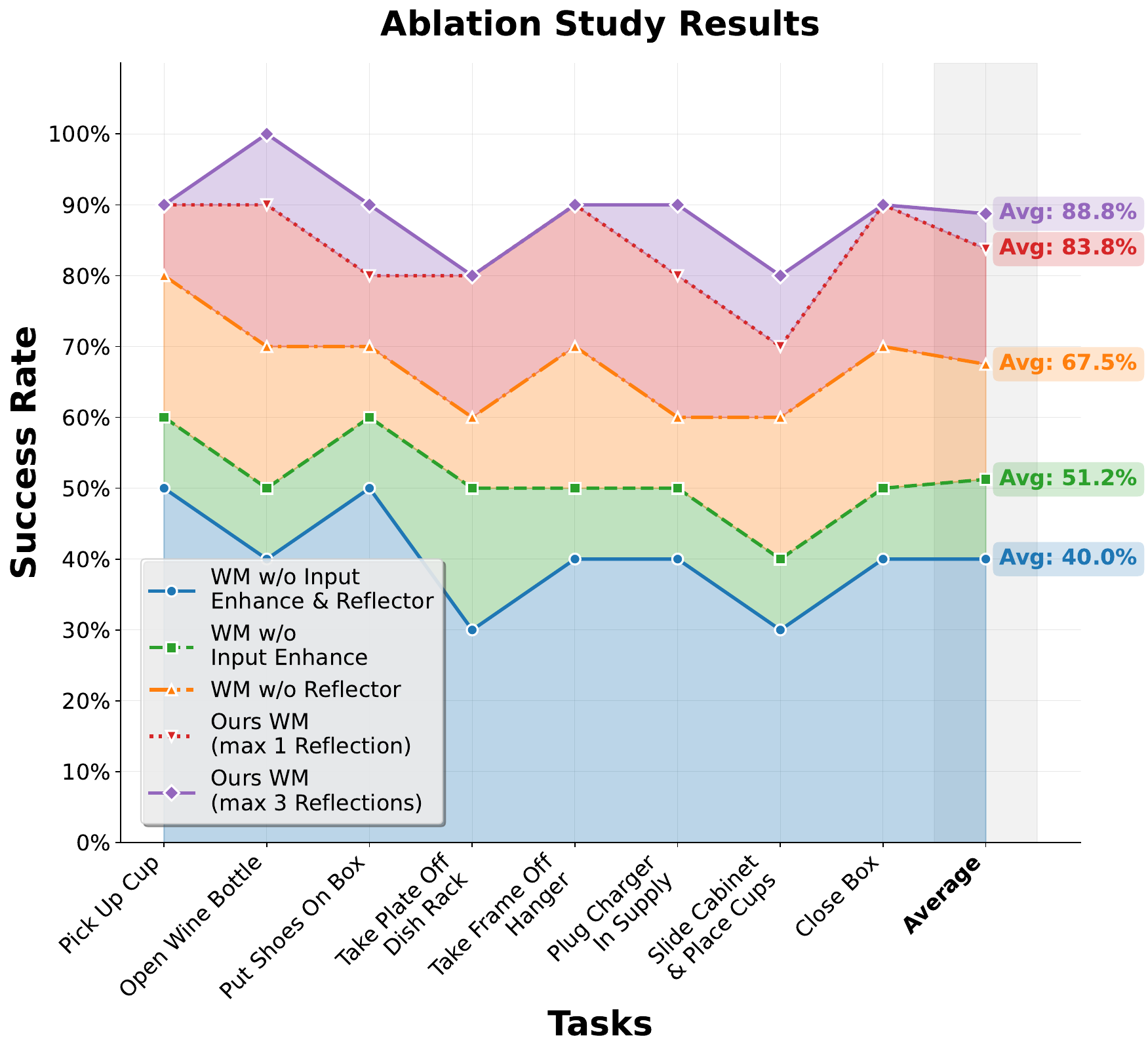}
    \caption{Ablation Study. The performance of our full model
(”World Model w/ Instruction \& max 3 Reflection”), shown
by the purple line, surpasses all ablated variants.}
    \vspace{-10pt}
    \label{fig:ablation}
\end{figure}

\textbf{Analysis:} The results in Figure~\ref{fig:ablation} quantify the impact of each component. Starting from a 40.0\% success rate for the baseline model, adding Input Enhancement provides the most significant single improvement (+27.5pp), while the Reflector alone yields a substantial gain (+11.2pp). Our complete model, which combines both, performs best, and its success can be further boosted from 83.8\% to 88.8\% by increasing the maximum reflection iterations from 1 to 3. 

These results demonstrate that both components are critical and complementary, confirming their effectiveness and answering our second research question (Q2).

\subsection{Real World Experiments (Q3)}
Our framework is tested on four diverse real-world manipulation tasks to validate its practical applicability. As shown in Table~\ref{tab:real-world}, our method achieves a 60\% average success rate, significantly outperforming baselines like MOKA (22.5\%) and MolmoAct (27.5\%). Consistent with the simulation findings, the end-to-end model OpenVLA fails (0\%). These results further highlighting our generalization ability.

\begin{table}[b!]
  \centering
  \vspace{-10pt}
  \caption{Real World Experiments}
  \vspace{-2pt}
  \label{tab:real-world}
  \resizebox{\columnwidth}{!}{%
  \begin{tabular}{lccccc}
    \toprule
    Method      & \makecell{Tomato\\Placement} & \makecell{Table\\Sweeping} &  \makecell{Weighing \\Duck}  & \makecell{Bottle\\Stand-Up}  & \makecell{Average \\Success Rate} \\
    \midrule
    OpenVLA   & 0/10 & 0/10 & 0/10 & 0/10 & 0\% \\
    MOKA        & 5/10 & 1/10 & 3/10 & 0/10 & 22.5\%    \\
    MolmoAct    & 5/10 & 0/10 & 6/10 & 0/10 & 27.5\%    \\
    \rowcolor{customblue}\textbf{Ours}      & 9/10 & 4/10 & 7/10 & 4/10 &  60\%      \\
    \bottomrule
  \end{tabular}%
  }
  % \vspace{-5pt}
\end{table}

\begin{figure}[t!]
    \centering
    \includegraphics[width=\columnwidth]{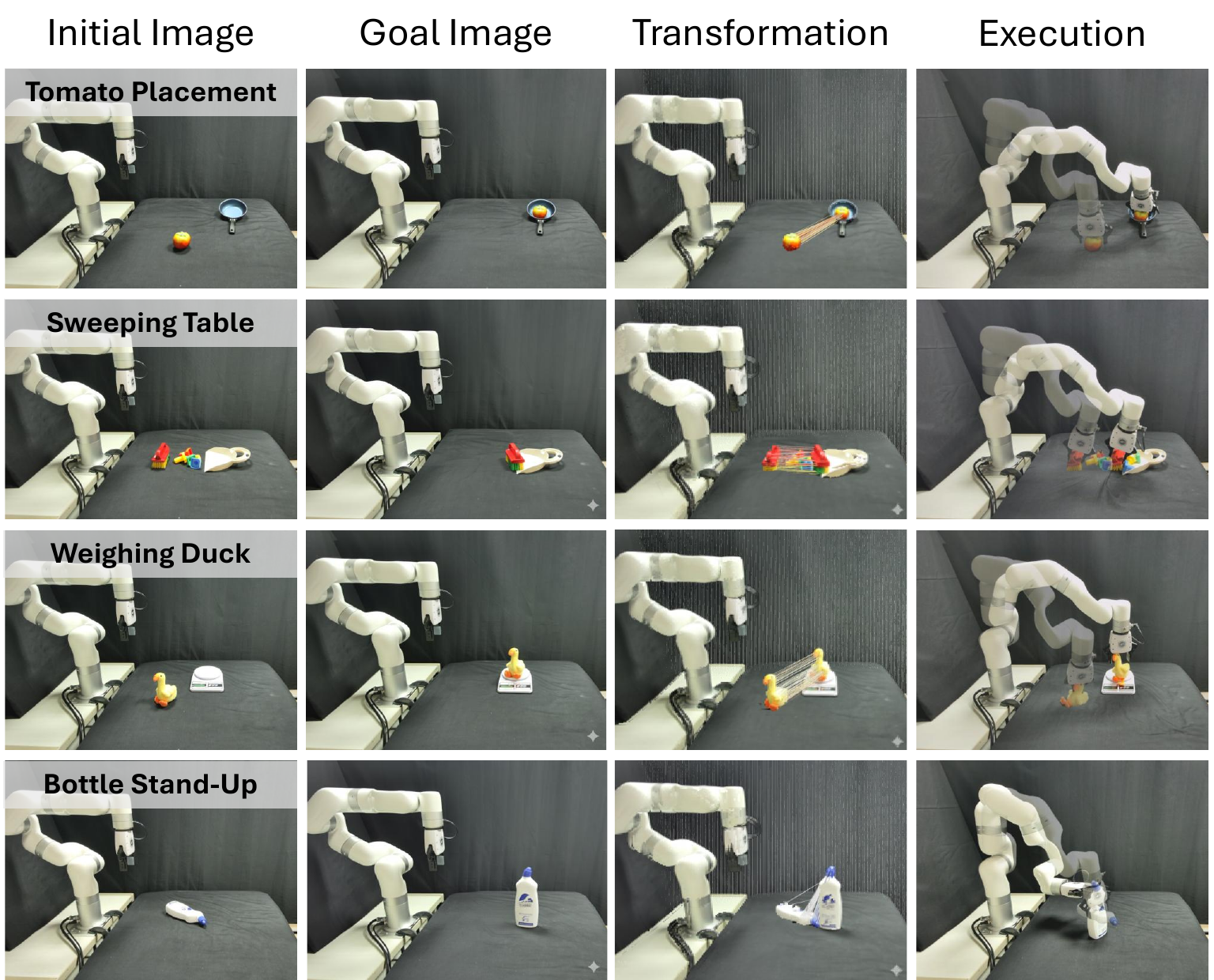}
    \caption{Visualizations of Real-World Experiments.}
    \label{fig:real world}
\end{figure}

Figure~\ref{fig:real world} provides qualitative evidence for these findings. The visualizations illustrate how the generated goal image captures the task's intent, and the subsequent transformation provides a precise spatial target, enabling the robot to execute complex tasks in a robust, zero-shot manner.

The framework's success across diverse tasks, objects, and environments (simulated and real), combined with its zero-shot deployment on different robot embodiments, demonstrates strong generalization, providing a clear answer to Q3.

\subsection{Failure Cases Analysis}
In our real-world experiments, we observe several typical failure modes as different tasks place varying demands on each module of our framework.

Failures originating from the Spatial Grounding module are the primary obstacle in several precision-demanding tasks. For \textit{Tomato Placement} and \textit{Weighing Duck}, this manifests as inaccuracies in the estimated depth, with the latter task being sensitive to minor height errors. Similarly, the \textit{Bottle Stand-Up} task's success is contingent on orientation precision, where even minor inaccuracies in the computed orientation from this module may lead to failure.

In contrast, the primary difficulty in the \textit{Table Sweeping} task shifts to the High-level Reasoning module. Generating a goal image representing a practical and feasible sweeping motion demands a sophisticated semantic understanding, and occasional failures occur at this reasoning stage.

\section{Conclusion}
\label{Conclusion}

In this work, we introduce \textbf{\name}, a hierarchical manipulation framework designed to address the critical challenge of zero-shot generalization. Our key contribution is a decoupled architecture where a high-level generative VLM, acting as an object-centric world model, envisions a visual goal state. This visual goal is iteratively refined for plausibility and accuracy through our novel Reflection-through-Synthesis mechanism. Crucially, this visual representation is translated into a precise 3D object pose, which serves as an explicit, training-free guide for a separate low-level policy. This design ensures the entire manipulation process, from semantic reasoning to physical execution, is performed in a zero-shot setting, eliminating any requirement for action data or task-specific finetuning. Our extensive evaluations in both simulation and the real world demonstrate that \name significantly surpasses existing end-to-end and hierarchical baselines across various manipulation tasks, highlighting its strong generalization and cross-embodiment capabilities.

% A key avenue for future work is to enhance the semantic consistency of the entire pipeline. Currently, we rely on separate, specialized models (e.g., Grounded-SAM, Depth-Anything) to produce masks and depth maps from the generated goal image. A critical weakness is that these downstream models are agnostic to the initial text instruction, which can cause semantic misalignments. A promising direction is to develop a unified high-level model trained to directly output a complete visual goal state—including the RGB image, object mask, and depth map—all conditioned on the language instruction. Such an end-to-end generative approach would ensure all components of the goal representation are semantically aligned with the user's intent, improving the framework's overall robustness.

% \addtolength{\textheight}{-12cm} 
\bibliographystyle{IEEEtran}
\bibliography{IEEEabrv,ref}

\newpage
\appendix
\subsection{Acknowledgements of the Use of AI}

The research presented in this paper incorporates several publicly available foundation models as components within our framework. These models are utilized for various functions, including generative image synthesis, semantic evaluation, object segmentation, and depth generation.

In addition to their role as research components, AI-powered tools assist in the preparation of this manuscript and its associated code. We use a large language model (Gemini, Google; accessed Aug–Sep 2025) for grammar correction and to improve text clarity. AI-assisted coding tools, such as Cursor, are employed during development. All code generated with AI assistance is manually reviewed and rigorously tested by the authors.

The authors take full responsibility for the content and conclusions of this work.

\subsection{Simulation Task Description}
\label{Simulation Task Description}
The simulation experiments are conducted on a suite of eight challenging manipulation tasks from RLBench. These tasks include: \textit{Pick Up Cup}, which involves grasping a cup and lifting it; \textit{Open Wine Bottle}, which requires the gripper to directly grasp the bottle cap and pull it off the bottle's opening; \textit{Put Shoe On Box}, picking up a shoe and placing it onto a box; \textit{Take Plate Off Dish Rack}, lifting a plate from a rack; \textit{Take Frame Off Hanger}, lifting the picture frame along the contour of the hanger to disengage it; \textit{Plug Charger In Supply}, plugging a charger into a power socket; \textit{Place Cup on Cabinet}, picking up a cup and setting it onto a cabinet; and \textit{Close Box}, manipulating a lid to close a box. For the "Pick Up" and "Take...Off" tasks, success is achieved as soon as the object is lifted clear from its supporting surface (e.g., the table or rack). In contrast, the "Put...On" and "Place...on" tasks require the robot to first successfully grasp the target object and then transport it to the specified goal location before releasing it.

\subsection{Real World Task Description}
\label{Real World Task Description}
The \textit{Tomato Placement} task requires the robot to pick up a tomato from its initial location and place it inside a pan, where the tomato must rest stably within the pan without rolling out.
In the  \textit{Table Sweeping} task, the robot should utilize a brush to sweep several screws scattered on a tabletop into a designated dustpan.
The \textit{Weighing Duck} task requires the robot to pick up a toy duck and place it accurately onto a weighing scale's platform such that it remains stable for a consistent weight reading.
\textit{Bottle Stand-Up} is a re-orientation task where the robot must grasp a bottle lying horizontally and set it in a stable, upright position.

\end{document}